# Explainable offline automatic signature verifier to support forensichandwriting examiners

Moises Diaz · Miguel A. Ferrer · Gennaro Vessio

## Abstract

Signature verification is a critical task in many applications, including forensic science, legal judgments, and financial markets. However, current signature verification systems are often difficult to explain, which can limit their acceptance in these applications. In this paper, we propose a novel explainable offline automatic signature verifier (ASV) to support forensic handwriting examiners. Our ASV is based on a universal background model (UBM) constructed from offline signature images. It allows us to assign a questioned signature to the UBM and to a reference set of known signatures usingsimple distance measures. This makes it possible to explain the verifier's decision in a way that is understandable to non- experts. We evaluated our ASV on publicly available databases and found that it achieves competitive performance with state-of-the-art ASVs, even when challenging 1 versus 1 comparison are considered. Our results demonstrate that it is possible to develop an explainable ASV that is also competitive in terms of performance. We believe that our ASV has the potential to improve the acceptance of signature verification in critical applications such as forensic science and legal judgments.

## Abbreviations

| | |
|---|---|
| ASV | Automatic signature verification |
| FHEs | Forensic handwriting examiners |
| UBM | Universal background  model |
| $LR_q$ | Likelihood ratio |
| $P(U)$ | Probability of membership in the UBM |
| $P(R)$ | Probability of membership in the reference set |
| DL | Deep learning |
| FCN | Fully convolutional network |
| DET | Detection error trade-off |
| FAR | False acceptance rate |
| FRR | False rejection rate |
| RF | Random forgery |
| SF | Skilled forgery |
| EER | Equal error rate |

## 1.Introduction

Forensic Handwriting Examiners (FHEs) are tasked with comparing a questioned signature to one or more genuine specimens. This practice is based on the assumption that it is highly unlikely for two different writers to produce identical handwriting. However, all writers exhibit inherentvariability in their writing or signature. Despite FHEs relying on established best practices and accumulating experience over time, their traditional examination methodology often faces criticism. This criticism stemsfrom concerns about the reproducibility and validity  oftheir visually based assessments, which inherently possess subjective elements. In other words, two experts can arrive at different conclusions. Therefore, there is a pressing need for an objective and quantifiable methodology in the field of forensic signature analysis, one that seeks to establish new standards and introduce scientific rigor into the process.

The pattern recognition and computer vision community have developed several automatic signature verifiers (ASVs). As a result, outstanding performances have been achieved on several benchmark datasets. In particular, exceptional results have been possible thanks to sophisticated and effective deep learning models that comprise hundreds of layers and millions of parameters [16, 19]. Nevertheless, performance is not the only attribute for signature verification systems. In addition to the seven features identified by Jain for biometric systems [33] and other pertinent aspects like security, fairness, and privacy [34], explainability has emerged as a crucial factor for the practical implementation of these systems [5]. The lack of explanation in the functioning of these systems prevents their use in some practical contexts, such as finance, healthcare, government, commercial transactions, and security. An easy-to-explain and straightforward computer system with principles close to the practice of forensic examination is preferable for a jury [44].

The trade-off between the explainability and security of ASV systems presents a multifaceted consideration. While transparent and interpretable features and classifiers enhance comprehensibility, they also introduce potential vulnerabilities that adversaries could exploit in forging signatures, as emphasized by [28]. Striking the right balance necessitates meticulous evaluation, possibly entailing the amalgamation of transparent features for typical scenarios and the incorporation of additional, less transparent security layers for safeguarding sensitive applications. The imperative for regular updates and continuous research cannot be overstated in mitigating evolving threats. Ultimately, the decision regarding the trade-off among explainability, security, and performance hinges upon the specific use case and the risk tolerance of the deploying organization.

Greater transparency should be considered to justify the use of an ASV system in a courtroom: The automatic extraction of parameters from a signature should be directly linked to its physical shape. So, a matching technique that is easy to interpret and explain in an objective and understandable way is needed. Moreover, there is a requirement for a measure to weigh the evidence of a conclusion drawn with simple, transparent, meaningful, and understandable scores. In summary, there is a practical use for a signature system capable of being explained in a human-to-human interaction [52]. Such a system would provide the forensic expert not only with a powerful tool to make the findings less questionable, but also a way to add absolute transparency in communicating the value of evidence, which would also ensure reproducibility and convince authorities to use biometric technology to explain the decision-making processes.

To this end, we propose an offline ASV inspired by the new technological paradigm of explainable artificial intelligence [52], which shows desirable features for FHEs. The characteristics of our system and its analysis are detailed below:

-It is designed in a modular way to improve its explicability [52]. Thus, at each step in the pipeline (i.e., for each module), the proposed techniques can be easily changed according to their accuracy and/or transparency.

-The ASV uses explainable features, which have a physical meaning close to the structure of the signature. Furthermore, our experiments consider exploring the performance with non-explicable deep learning, DL, features for performance comparison purposes. Understandable matching distances are studied to match signatures.

-We select the most straightforward ones to make the meaning clear across all available distances.

-Forensic analysis requires a transparent model of the "world." In our case, we use a universal background model (UBM) constructed with a third-party set of signatures. Synthetic signatures are also studied in the UBM to improve the privacy of the system.

-The system is also quite robust in supporting 1 versus 1 signature comparisons, representing the worst working scenario for an FHE.

-The results are provided in terms of evidence, i.e., likelihood ratio, LR, and probability of belonging to the UBM and to the reference population, which are criteria familiar to forensic experts.

The results validate the use of explainable and DL features in several situations. Also, we assess the use of a real versus synthetic UBM. We selected several publicly available offline signature databases for these purposes, whose acquisition process is also transparent. As for the synthetic UBM, the generation of synthetic signatures is based on motor control procedures, thus assuring its transparency. This also increases the data privacy property of the proposed explainable system. Finally, it is worth noting that while we are witnessing a paradigm shift in which signatures are increasingly acquired through online acquisition systems such as digitizing tablets and smartphones, FHEs are still required to verify signatures acquired primarily in *offline* mode. For this reason, we focus here on the latter case.

## 1.1. Subsequent work and main contribution

In the initial version of this work [18], an offline automatic signature verifier was proposed for forensic applications and validated on two public databases. We also analyzed individual handcrafted features to assess their impact on the verifier's performance.

However, subsequent reviews and publications have emphasized the need for quality assurance and the reliability of systems in legal contexts [2, 14, 39], highlighting a general lack of forensic handwriting examiners [14]. Taking these factors into consideration, we believe that developing explainable systems that can provide objective and reliable decisions in signature verification would be a valuable contribution to forensic handwriting analysis.

To this end, we have extended our previous work and proposed an explainable offline automatic signature verifier for forensic applications based on the verifier presented at the conference. Our experimentation now includes more databases and a more comprehensive comparative analysis between explainable and non-explainable signature verification systems. Our detailed analysis leads us to present this system as an explainable ASV.

The rest of this paper is organized as follows. First, we review related work in Sect. 2. Then, Sect. 3 introduces the proposed automatic system for forensics, which is set up in Sect. 4. Experimental evaluation and results are reported in Sect. 5, while the article is concluded in Sect. 6.

## 2. Related work

The demand for interpretable systems has primarily been tackled within the field of biometrics to elucidate the out- comes of deep neural network-based pattern recognition. One notable illustration is Grad-CAM, a widely recognized post-processing explanation technique [62], which functions as a visual aid. Its use has been demonstrated for ear recognition systems [3] as it can identify the most distinctive patterns in images, thus helping to provide textual explanations and even detecting the most discriminating neurons in the network. Similar instances of interpretable techniques are evident in iris recognition [10], fingerprint segmentation [36], and face recognition [73], all of which leverage the physical attributes of the subject [60].

All of these biometric recognizers are based on DL technology, which has improved explainability with visualization patterns. In offline signature verification, the rise of DL has also achieved impressive results on different benchmark datasets, e.g. [25, 26, 74]. For instance, Grad- CAM can identify regions of interest within an image of a signature that the model deems important to its decision, thereby assisting forensic experts in comprehending which parts of a signature played a crucial role in determining its authenticity. Nevertheless, our objective is not to elucidate why a neural network-based model classified a signature as genuine or forged, but rather to create a system that is inherently easy to explain. Although Grad-CAM can offer insights into the model's decision-making process, neural networks are still perceived as complex models, and conveying their understanding and explanation to a court necessitates a background unfamiliar to most forensic examiners. Furthermore, to effectively utilize neural net- works and, by extension, Grad-CAM, access to a substantial amount of labeled data is essential, which is not always the case in a typical work scenario where only a few signatures are accessible. To the best of our knowledge, a fully explainable biometric system in the context of offline ASV, up to the development of our system, has not been available.

Traditionally, to examine signatures, FHEs use various illumination and magnification tools, such as stereo-microscopes, light panels, specialized grids printed on trans- parent films, and so on. The comparison results are then typically provided on a five or nine-point scale, based on some hypotheses about the genuineness of the questioned signature [68].

The literature is sparse on systems specifically oriented to supporting forensic investigations. Some automated tools have been developed in recent years to support the daily activity of FHEs, such as FLASH ID, iFOX, and D-Scribe [16]. However, FHEs have traditionally made very limited use of such systems. Additionally, some tools have some drawbacks that affect the possibility of their usein real-world forensic cases. For example, in a courtroom, acareful understanding and explanation of the decision process are crucial [69]. Ultimately, none of these systems is so widespread today that its usefulness can be fullyestablished.

As an illustration, the use of signatures is prevalent inbank checks [23]. The verification of authorship can beopen to interpretation by an FHE and may often requirejustification in a court of law. The solution presented in thisarticle offers evidence of the authorship of signaturesthrough the use of likelihood ratios as metrics. Further-more, the transparency of our offline ASV system opensthe door to the acceptance of the system by judges andother professionals with limited technological knowledge.

To partially overcome this problem, and to provide empirical evidence on the reliability of decisions made by computer systems, competitions were organized by bio-metric researchers, in collaboration with forensic examiners, to evaluate the performance of automated tools inforensic cases Popular competitions include the 4NsigComp, SigComp, SigWiComp series [7, 41, 42, 48, 49], in which sub-corpora closest to the real forensic cases were used, and organizers presented the results in terms of LR. The ultimate objective in these competitions was performance. These competitions did not relate to the explainability of the ASVs nor was the use considered of one signature as a reference.

To try to bridge the gap between biometrics and forensics, M. I. Malik's thesis [47] proposed new signature verification systems based on local features and on the assumption that local information contains essential clues for clear explanation rather than taking an holistic approach. Similarly, Marcelli et al. [50] proposed a method to link the features FHEs are most familiar with and measurements on the digital image of a signature on a paper document. The method was intended to provide a quantitative estimate of the variability of features when examined in different contexts. More recently, Okawa [55] has proposed an approach to mimic the cognitive processes that FHEs use to reach their decision, based on a ''bag-of visual words'' and a vector of locally aggregated descriptors. The main idea is to focus on the salient local regions of the signatures in order to conduct the comparisons. Inspired by these works, we also use local features to represent the signatures.

By contrasting them with relevant published work on offline automatic signature verification systems, Table 1 provides several techniques and algorithms. It is important to note that the explainability of various systems varies. However, it is outside the purview of this study to categorize these systems in terms of their explainability. Instead, we concentrated on developing a method that could be useful to forensic handwriting experts.

In this paper, we propose a different approach. In addition to comparing a questioned signature with a reference set, we compare such a signature with a UBM developed with external signatures. This approach is similar to a police procedure, in which the police have their own UBM to compare the cases they receive. The use of a UBM is not new in signature verification. We identify a few previous works that used a trained UBM in online signature verifiers. For example, a Gaussian Mixture Model was introduced with features of all signatures in a UBM in [51, 75]. Instead, an Ergodic-HMM was preferred to train a UBM in [4]. The signature system used the trained UBMtosupport the final decision in these cases. However, these trained UBMs fail in their explicability since they are too graphically opaque for human users to understand [28].

Our explainable ASV is designed with several modules of features and classifiers. These modules can be cus tomized to meet the specific needs of different applications, as some applications require a higher level of explainability than others. Additionally, our work explicitly considers the security and forensic requirements of ASV, such as the need for robustness to attacks and the need to generate explainable predictions. Finally, our ASV has been evaluated on large and diverse signature databases, demonstrating its effectiveness.

## 3. Explainable ASV architecture

The architecture of our system aims to verify a questioned offline signature automatically and is composed of several modules, as illustrated in Fig. 1. Each module is a step in a pipeline that can be implemented with different techniques to analyze their impact on the final performance. Additionally, each module can be easily replaced depending on the desired level of explainability in an end application or for a user.

Our system requires a reference set composed of enrolled signatures to the ASV, i.e., real reference signatures of individuals. Additionally, a transparent UBM, which ought to be understood by itself [40], is used as input. The UBM is a gathering of representative sample from the population. Toward an explainable system, we propose to use a pool of signatures where the features are extracted and compared directly with the questioned and reference set. It is worth highlighting that the UBM is challenging to build, as it should represent all signature populations except for the reference set. A practical solution is to build up this set with signatures signed by other people. To this end, we used publicly offline signatures databases. Ideally, this UBM should be independent of the written script.

The output of the system numerically estimates whether the questioned signature is genuine or not through the probabilities of belonging to the reference set and the UBM, and the LR. Rather than a simple numerical score, the LR estimates from the evidence whether a questioned specimen is closer to the reference signatures or to the UBM [11]. Furthermore, the design of the verifier uses an explainable framework with features and matching mechanisms that are more understandable to human beings. In summary, the hypothesis behind this system is that a genuine signature must be closer to the reference set than the UBM set; and a fake specimen must be farther from the reference set than the UBM.

*Table 1 Offline automatic signature verification techniques for forensic handwriting examination*

| Algorithm/technique | Description | References |
|---|---|---|
| Graphology analysis | Examines psychological aspects of handwriting to assess authenticity | [56] |
| Machine learning | Utilizes supervised learning models for signature verification based on training data | [23] |
| Local features | Identify keypoints or interest points in signatures for robust feature matching. | [47] |
| Stroke sequence analysis | Studies the order and sequence of strokes in a signature | [50] |
| Local binary patterns | Encodes texture information for signature representation and comparison | [53] |
| Handwriting feature analysis | Analyzes individual handwriting features like slant, size, and pressure | [68] |
| Graph-based approaches | Represent signatures as graphs and analyze structural information for authentication | [45] |
| Forensic document analysis | Investigates paper, ink, and other physical aspects of the document | [55, 69] |
| Signature comparison | Compares questioned signatures to known reference signatures | [17] |
| Biometric authentication | Uses biometric data like pen pressure and speed for verification | [19] |
| Neural networks | Employs deep learning models for complex signature analysis | [43, 74] |

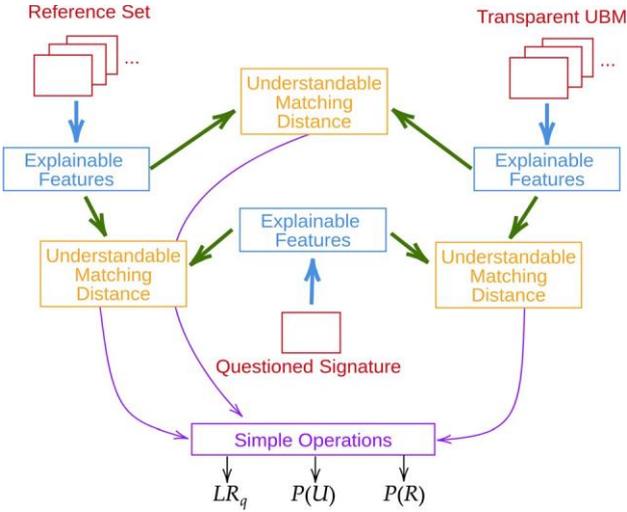

*Fig. 1 Block diagram illustrating our explainable offline ASV system. The inputs consist of the reference set containing known signatures, a transparent Universal Background Model (UBM) constructed using a third-party set of signatures, and questioned signatures that have never been encountered by the ASV. The outputs are presented in terms of evidence, specifically the likelihood ratio (LRq), as well as the probabilities of membership in the UBM and the reference set ([P(U), P®])*

$x=[x_1, x_2,…,x_m]$. Thus, our system proceeds with specimens interpreted as features. Accordingly, feature extraction is directly carried out with all signatures in the UBM, the reference signatures, and the questioned one. To achieve interpretability, the features used should be easily understood by a human being and related to physical or geometrical characteristics of the signatures. To this aim, explainable handcrafted features are used for pursuing model explainability. For the sake of transparency, we select some of the well-known explainable features such as geometrical relations, curvature properties and slants.

Let $r \in R$ be a reference specimen and R the reference set. Also, let U be the total number of signatures included in the UBM and u a single signature of the UBM. Finally, we define q as a questioned signature to be verified automatically. Next, the explainable ASV calculates an LR and two types of probabilities.

### 3.1 Likelihood ratio

We compute the likelihood ratio, LRq, for the questioned signature. It evidences how many times q is closer to the reference signatures than to the UBM. To compute LRq, we calculate two distances. A first distance, $\delta_1(q)$, represents the alternative hypothesis, H1. This distance quantifies how close a signature is to other signers, $u \in U$. Mathematically, it is denoted as follows:

$$\delta_1(q) = \min_{\forall u \in U} d(q, u) \quad (1)$$

where $d\,(\cdot)$ denotes a generic distance matching between the questioned signature and a signature belonging to the UBM, $u \in U$. We express the null hypothesis, $H_0$, by measuring the distance between the questioned signature and the available reference signatures, $r \in R$, belonging to the claimed signer. The following relation is performed to compute $\delta_2(u)$:

$$\delta_2(q) = \min_{\forall u \in r} d(q, r) \quad (2)$$

Finally, the evidence is computed in terms of LR(q) as:

$$LR_q = -2 \log \left( \frac{\delta_2(q)}{\delta_1(q)} \right) \quad (3)$$

which means how many times the questioned signature is closer to the reference signatures than to any other signature. New scores in terms of probabilities are worked out to improve the interpretability of such a ratio.

*3.2 Probabilities*

In addition to LRq, the system calculates two probabilities, [P(U), P(R)], which can provide more transparency to the verification. P(U) denotes the probability of belonging to the UBM. Alternatively, P(R) gives the probability of belonging to the reference set. While the first probability can always be computed, the second requires more than one signature in the reference set in order to be estimated. It is, therefore, expected that the more reference signatures are available, the better the estimation of P(R). As such, our offline ASV cannot compute P(R) in the case of 1 versus 1 signature verification.

3.2.1 Probability of belonging to the UBM

The metric P(U) represents the probability that the questioned LR belongs to the UBM. It can be computed as follows:

$$P(U) = 1 - F_{LR_{u_i}} = \left( LR_q | \mu_{LR_{u_i}}; \sigma_{LR_{u_i}} \right) \quad (4)$$

Here, $F_{LR_{u_i}}$ is the normal cumulative distribution function (CDF) with a mean and standard deviation $\left( \mu_{LR_{u_i}}; \sigma_{LR_{u_i}} \right)$, evaluated at the questioned LRq. The graphical represen tation of P(U) is the area under the normal probability density function (PDF), $F_{LR_{u_i}}$ in the interval [LR$_{q, +\infty}$]. To obtain this, we need to compute the LR of each specimen in the UBM, LR$_u$ :

$$LR_{u_i} = -2 \log \left( \frac{\delta_{U,2}(u_i)}{\delta_{U,1}(u_i)} \right) \quad (5)$$

where the signature i in the UBM is denoted by $u_i$ and ($\delta_{U,1}$, $\delta_{U,2}$) are two distances that depend on the UBM and the reference set, respectively. These two distances can be estimated as follows:

$$\delta_{U,1}(u_i) = \min_{\forall u \in U} d_{u \backslash u_i}(u_i, u) \quad (6)$$
$$\delta_{U,2}(u_i) = \min_{\forall u \in R} d_{u \backslash u_i}(u_i, r) \quad (7)$$

d(·) being a matching distance of two signatures in the form of feature vectors.

### 3.2.2 Probability of belonging to the reference set

This probability, P(R), assesses how probable it is that the questioned signature may belong to the reference specimens. The following expression can be used to calculate it:

$$P(U) = 1 - F_{LR_{r_i}} = \left( LR_q | \mu_{LR_{r_i}}; \sigma_{LR_{r_i}} \right) \quad (8)$$

In this case, P(R) represents the area under the $F_{LR_{r_i}}$ curve where LRq is likely to fall within the interval [-∞, LRq.].. Thus, the LRq is evaluated in the CDF, $F_{LR_{r_i}}$ with a mean and standard deviation $\left( \mu_{LR_{u_i}}; \sigma_{LR_{u_i}} \right)$ . Similarly, we can determine the LR of each reference signature as follows:

$$LR_{r_i} = -2 \log \left( \frac{\delta_{R,2}(r_i)}{\delta_{R,1}(r_i)} \right) \qquad (9)$$

where $(\delta_{R,1}(r_i), \delta_{R,2}(r_i))$ are two distances that depend on references and UBM specimens:

$$\delta_{R,1}(r_i) = \min_{\forall u \in R} \ d_{r \setminus r_i}(r_i, r) \quad (10)$$

$$\delta_{R,2}(r_i) = \min_{\forall u \in U} d(r_i, u) \qquad (11)$$

The above procedure aims to describe a transparent system which is generic, not only on the feature space but also on distance matching. A visual easy-to-understand example of outputs generated by our system can be viewed in Fig. 2.

## 4 Setting up the proposed system

We used two different databases to develop the UBM. We also used three other databases for the experiments to avoid bias in the results. This strategy contributes significantly to the ASV explainability [5]. All used signatures are also transparent in terms of the acquisition protocol and composition.

Two UBMs were developed, one with real signatures and another with synthetic ones. The computational model to generate the synthetic signatures was based on well known control motor processes, and the procedure is completely transparent [22]. It provides complete privacy to the explainability of the UBM because none of the signatures could be identified. The former was created with the first n genuine signatures of GPDS960 [72], whereas the latter with the Synthetic10000 [22]. The motivation is to study our system in these two situations, the actual use of which would depend on the real application. The experiments were conducted with offline specimens available in the MCYT-75 [57], BiosecurID [24], Thai [12], and CEDAR [37] databases. Please note that all these databases are publicly available and allow experiments using two scripts.

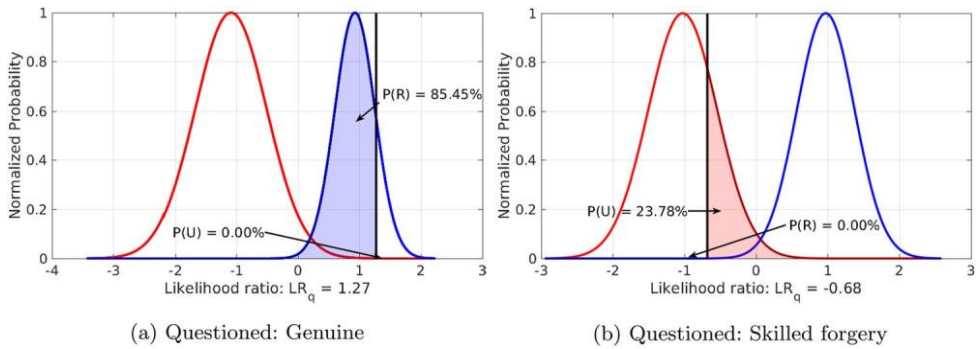

(a) Questioned: Genuine      (b) Questioned: Skilled forgery

*Fig. 2 An interpretable visualization of the output results of the proposed ASV system is presented, showcasing scenarios where questioned signatures are either a genuine (on the left) or a skilled forgery (on the right). The normalized probabilities are depicted as functions of the likelihood ratios. In both figures, the solid red and blue curves represent the normalized probability density functions of the UBM and the reference sets, respectively, denoted as $F_{LR_{ru_i}}$ and $F_{LR_{r_i}}$. As expected, when the questioned signature is genuine, the probability of belonging to the UBM (P(U)) is lower, while the probability of belonging to the reference set (P(R)) is higher.*

### 4.2 Features

We used four types of handcrafted features in our proposal as explainable features. DL-based features were also considered to quantify the performance for each feature. Explainable features could be accepted in a court if they were close to the physical features made by ink deposition on paper, as non-expert humans may then understand how they were obtained. Moreover, the value of the parameters is related to the physical phenomenon of ink deposition of the signature. On the other hand, DL features can perform better [28]. The features utilized in this paper can be shared upon request.

As for explainable features, the first type is based on several geometrical descriptions of the shape of the signature. We work out these measures in Cartesian and polar coordinates based on their contours [20]. All geometrical features were concatenated into a vector, resulting in a dimension of 445. The second type consisted in dividing the signatures recursively into a quadtree at two levels. Then we computed the gradient on each level to explain the texture properties of the signatures [63]. The dimension was 200. The third type was based on counting the run lengths of the binary images. The feature vector dimension was 400, and it explains the width of the strokes in the four main directions: vertical, horizontal, diagonal 45º, and diagonal 135º [9]. The last feature type consisted in determining the physical textural properties of offline signatures [21]. Specifically, we elaborated the local binary patterns and local derivative patterns of the images. The dimension of these vectors is 765 and 255, respectively. Moreover, we applied the discrete cosine transform to these vectors, obtaining dimensions of 168 and 167, respectively. As a result, four feature vectors associated with textural features were obtained. These features are summarized in Table 2.

There are two main advantages in using these features. The first is that they have been successfully applied to offline ASVs in previous work and some forensic-based signature competitions such as 4Nsig-Comp2010 [7] and SigWIcomp2015 [49]. The second is that the LBP, LDP, and gradient features, organized in quadtree structures, quantify how the ink was deposited on the paper from different points of view. Run-length and geometrical features work on binary images. Therefore, they mainly reflect aspects related to the shape of the signature from different perspectives. Furthermore, as the extraction process is entirely different in each case, we take advantage of the complementarity of their information content.

As for the DL features, three deep neural networks were involved. The first two models consisted of popular convolutional neural networks, namely VGG19 [66] and ResNet_v2 [31], with weights pre-trained on ImageNet. To perform feature extraction, we followed the common practice of removing the top-level classifier, adding a global average pooling on top of the convolutional base, and performing transfer learning. Additionally, we con ducted transfer learning not from ImageNet but from a knowledge base more similar to our context. To this end, we also used an ad hoc, fully convolutional network (FCN) with weights we have pre-trained on the very popular MNIST dataset of handwritten digits. Having no fully connected layers, this model can accept inputs of any size, so it is suitable for processing small MNIST digits and higher resolution signature images. We applied commonly suggested ''generic'' (VGG19-g) pre-processing for all three models, which consists of scaling the signatures to 224 224 pixels and normalizing their values between 0 and 1. In this way, the global average pooling provides 512features from VGG, 2048 for ResNet_v2, further reduced to 512 with a simple auto-encoder, and a smaller 64-dimensional feature vector from the simpler FCN. Also, to experiment with a more ''specific'' (VGG19-s) pre-processing for signature images (as done for example in [15]), we resized the images to 155 220 pixels (to preserve the aspect ratio), inverted the pixel values so that the back ground pixels have value 0 and finally normalizing these values between 0 and 1. For simplicity, we have applied this specific pre-processing only with VGG19.

It is worth noting that our goal was to assess whether the explainable features are competitive enough, when compared with automatically learned features, without much loss of performance. For this reason, we used the common practice of feature extraction with transfer learning without training much more specialized DL-based signature verification systems. Indeed, developing such a system was not the aim of our research. At the end of the experimental section, a comparison with more advanced methods is reported.

### 4.3 Matching distance

Many powerful offline ASVs use machine learning classifiers. However, we propose to use understandable matching distances for function-based features as they offer a more straightforward explanation for our ASV. In the literature, there exist matching distances for function-based features of different or equal length, such as the '2 norm. In our case, our matching will be based on features of the same size. We experimented with three distances for their extensive use in signature verification [16].

We decided to use a simple version of DTW. This matching distance builds a dissimilarity matrix with the Euclidean distances between all the members of two feature sequences to calculate the optimal distance between the elements of the two feature vectors. Excellent results have been achieved in offline and online signature verify- cation when this distance was used, e.g. [64]. Manhattan distance or '1 norm has also been used because of its good results in signature verification [61]. Finally, cosine distance has also been chosen for our experiments since it is based on the simple Euclidean dot product formula. It has proven its effectiveness via DL features over others, such as the Euclidean distance [58].

Motivated by their simplicity and effectiveness in sig nature verification, we integrated these methods into the proposed ASV. In addition, their use would help our aim to develop an explainable automatic signature verifier for offline signatures.

### 4.4 Evaluation

As usual in biometric systems and automatic signature verification, performance was evaluated through detection error trade-off (DET) curves. We computed the false acceptance rate (FAR) and false rejection rate (FRR) curves for this purpose. First, we used the signatures in the reference set and the questioned signatures to develop the FAR curve. To allow a fair comparison, the reference signatures were the initial signatures enrolled per user. Subsequently, the remaining genuine signatures were employed as questioned signatures. Then, two FAR curves were built for random and skilled forgeries. The random forgery (RF) set was worked out with a random signature out of 74 random users, whereas all fake signatures were used for the skilled forgery (SF) set. To quantify the per-formance in the verification task, the equal error rate (EER) was used in all cases.

It is noteworthy that various metrics have been proposed for the assessment of automatic signature verification systems. These metrics encompass the FAR, FRR, and Half Total Error Rate (HTER), as detailed in [6]. Furthermore, a recent standard ISO Central Secretary [32] promotes the adoption of the Attack Presentation

Classification Error Rate (APCER) at fixed Bona Fide Presentation Classification Error Rate (BPCER) values. In our research, we have chosen to uphold the universally accepted metrics [16, 59] to prevent potential confusion in the long-term development of the automatic signature verification field. It is important to note that these same metrics were employed in the most recent international ASV competition to evaluate system performance [70].

To evaluate our system, we used only one signature as a reference. Our motivation was to set up the system to cope with a very challenging case in ASV. Hence, it was expected that the more signatures are used in the reference population, the better the performance.

This evaluation allows us to study our proposed system forforensics in the context of other computational ASVs. Finally, the best configuration is evaluated by increasing the number of signatures in the reference set.

*Table 2  Feature vectors used in the explainable ASV*

| Explainable features | | | Deep learning features | | |
|---|---|---|---|---|---|
| Label | Features | Dimension | Label | Features | Dimension |
| g | Geometrical | 445 | d1 | VGG19-g. | 512 |
| qt | Quadtree | 200 | d2 | VGG19-s. | 512 |
| rl | Run-length | 400 | d3 | ResNet_v2 | 512 |
| t1-t4 | Textural | (765, 255, 168, 167) | d4 | FCN | 64 |

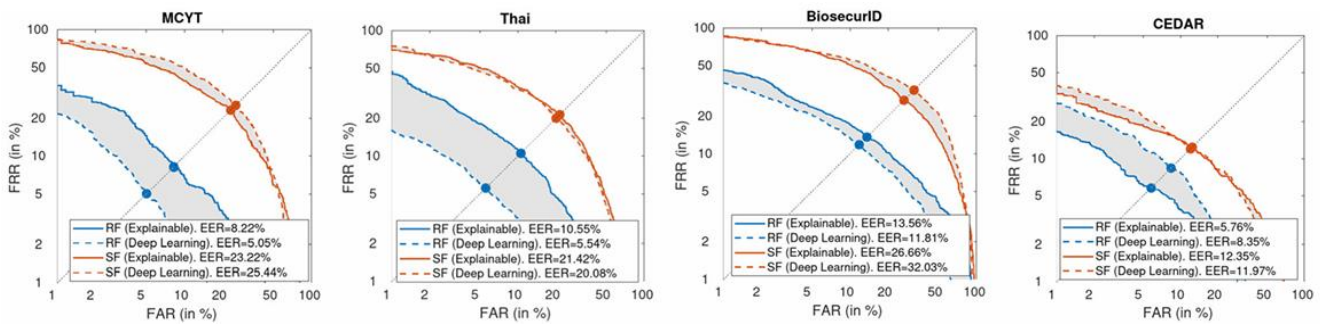

*Fig. 3 Detection error trade-off curves for the proposed ASV system, utilizing explainable features derived from both geometric and texturalproperties of the signature, as well as deep learning features extracted from VGG19 and ResNet_v2*

*Table 3 Equal error rates (%)*

| Database | Matching distance | Explainable features | Deep learn features | |
|---|---|---|---|---|
| MCYT | DTW | 8.22 | 5.05 | Random forgeries |
| | $\ell^1$ norm | 6.29 | 4.29 | |
| | Cosine | 8.94 | 4.09 | |
| Thai | DTW | 10.55 | 5.54 | |
| | $\ell^1$ norm | 7.81 | 4.76 | |
| | Cosine | 8.86 | 4.77 | |
| BiosecurID | DTW | 13.56 | 11.81 | |
| | $\ell^1$ norm | 8.16 | 9.85 | |
| | Cosine | 11.59 | 10.15 | |
| CEDAR | DTW | 5.76 | 8.35 | |
| | $\ell^1$ norm | 5.19 | 8.01 | |
| | Cosine | 7.17 | 6.50 | |
| MCYT | DTW | 23.22 | 25.44 | Skilled forgeries |
| | $\ell^1$ norm | 21.17 | 23.93 | |
| | Cosine | 21.89 | 24.47 | |
| Thai | DTW | 21.42 | 20.08 | |
| | $\ell^1$ norm | 20.58 | 20.25 | |
| | Cosine | 23.83 | 20.00 | |
| BiosecurID | DTW | 26.66 | 32.03 | |
| | $\ell^1$ norm | 26.03 | 31.08 | |
| | Cosine | 27.92 | 31.33 | |
| CEDAR | DTW | 12.35 | 11.97 | |
| | $\ell^1$ norm | 12.12 | 8.41 | |
| | Cosine | 13.18 | 11.14 | |

## 5. Experimental results

This section demonstrates that the proposed offline ASV is adequate for performance and explainability. We also studiedthe1versus1verificationcasetoadjust the ASV. Finally, the best configuration is evaluated by increasing the number of signatures in the reference set.

### 5.1. Studying explainable and deep learning features

Feature extraction is a crucial step in an offline ASV.

Indeed, many researchers have proposed several feature extractors to improve the performance of the systems[16]. However, in the previous decade,DLstrategies seemed to be the optimal technique for representing patterns in terms of performance [29]. In this section, we quantify the performance of our ASV system when explainable or DL features are used.

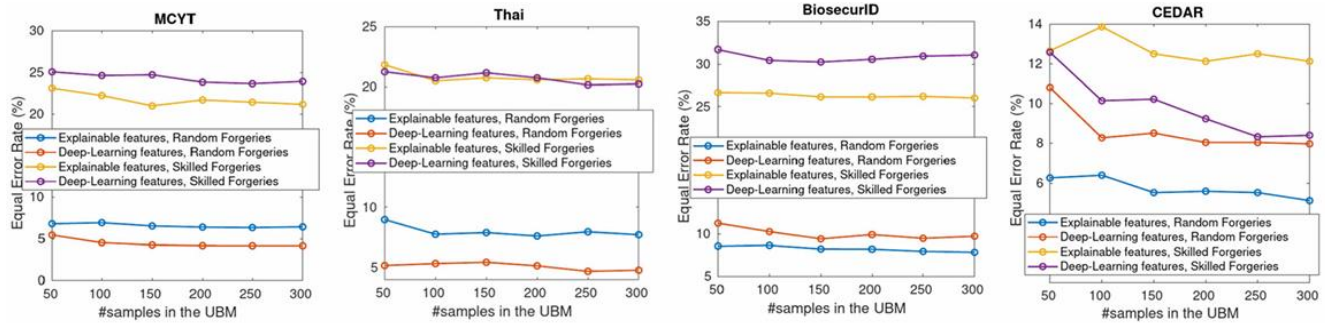

*Fig. 4 Exploring the optimal number of genuine signatures in the UBM across three databases. The experiments encompassed random andskilled forgeries, employing both explainable and deep learning features in each scenario.*

We set up the system using DTW as matching distance, 300 real signatures in the UBM, and a single reference signature. Besides using this setup in our previous work [18], training with one signature represents the most challenging case in signature verification [9]. To compute aglobal performance, we fused LRs and probabilities as if they were considered scores in a biometric-based signatureverifier. We then applied a weighted sum for the explain- able features (see Table 2) as follows: $s_{fused}^h = w_1\big(LR_g + P_g\big) + w_2(LR_{qt} + P_{qt}) + w_3(LR_{rl} + P_{rl}) + w_4(LR_{t_i} + P_{t_i})$ where the weights were proportionally adjusted, thus regarding the performance of each features:

$\Omega^h = (0.1; 0.75; 0.05; 0.1)$. For those who are interested, it may be helpful to consult the individual performance of hand-crafted features as presented in [18]. The same fusion strategy was applied to the DL features: $s_{fused}^{dl} = w_1(LR_{d1} + P_{d1}) + w_2(LR_{d2} + P_{d2}) + w_3(LR_{d3} + P_{d3}) + w_4(LR_{d4} + P_{d4})$ , the weights being $\Omega^{dl} = (0.5, 0.25, 0.15, 0.1)$.

Figure 3 shows the DET plots when explainable or DL features are used for each dataset and random, RF, and skilled forgeries, SF. For the MCYT corpus, a similar performance was obtained for SF with both features. The area between DET plots was 1.90. In the case of RF, slightly better performance was achieved with DL features with 3.02 of the difference between DET plots. These results are also consistent with the other two databases. We obtained 2.48 and 0.94 in DET curves for RF and SF in Thai, respectively. Similar differences were found in BiosecurID, with 2.30 and 4.36 for RF and SF. Thus, for RF and SF, the differences in the CEDAR case were 0.22 and 0.74. In the case of EER, Fig. 3 shows the performance obtained in each case. In general, using this initial con figuration of our system, explainable features offer similar results as DL ones, especially in SF, which represents the most challenging signature verification experiment.

*Table 4 Equal error rates (%) using real and synthetic genuine signatures in the universal background model using $\ell^1$ norm as matching distance.*

|  | Database | UBM | RF | SF |
|---|---|---|---|---|
| Explainable features | MCYT | Real | 6.29 | 21.17 |
|  |  | Synt. | 7.24 | 23.04 |
|  | Thai | Real | 7.81 | 20.58 |
|  |  | Synt. | 12.00 | 24.25 |
|  | BiosecurID | Real | 8.16 | 26.03 |
|  |  | Synt. | 8.33 | 25.27 |
|  | CEDAR | Real | 5.19 | 12.12 |
|  |  | Synt. | 7.00 | 11.82 |
| Deep learning features | MCYT | Real | 4.29 | 23.93 |
|  |  | Synt. | 4.99 | 22.42 |
|  | Thai | Real | 4.76 | 20.25 |
|  |  | Synt. | 6.81 | 22.92 |
|  | BiosecurID | Real | 9.85 | 31.08 |
|  |  | Synt. | 8.04 | 23.37 |
|  | CEDAR | Real | 8.01 | 8.41 |
|  |  | Synt. | 4.58 | 5.15 |

*5.2. Studying different understandable matching distances*

In this subsection, our goal is to analyze the performance of our system with DTW, $\ell^1$ norm and cosine, as understandable matching distances. Therefore, we kept the other setting options of our system unchanged, such as the use of real signatures in the UBM, one signature as a reference, and the results when the features are fused.

Table 3 summarizes the results we obtained for both random and skilled forgeries. We can compare the performance in terms of EER when explainable and DL fea- tures are used with each matching distance. It is observed that the lower loss of performance with explainable features is obtained with the $\ell^1$ norm. In the column ''Explainable features,'' we see that this distance out performs the other two. Additionally, we can compare the performance loss row by row with explainable and DL features. In terms of explainability, all distances support the concept of explainability in automatic signature verification. Among them, $\ell^1$ norm distance can be seen as the most useful in critical applications and the easiest to understand since it is based on the sum of the difference of individual elements in a vector. Here we will continue to explore the best system configuration with this matching distance.

*Table 5  Equal error rates (%)increasing the number of reference signatures*

| Database | UBM | Feature | Reference signatures | | | | | |
|----------|-----|---------|------|------|------|------|------|----------------|
| | | | 1 | 3 | 5 | 7 | 10 | |
| MCYT | Real | Explain. | 6.29 | 4.32 | 3.86 | 2.72 | 1.80 | Random forgeries |
| | | Deep L. | 4.29 | 2.52 | 1.73 | 1.03 | 0.74 | |
| | Synt. | Explain. | 7.24 | 4.43 | 3.57 | 2.88 | 1.82 | |
| | | Deep L. | 4.99 | 2.67 | 1.77 | 1.30 | 0.90 | |
| Thai | Real | Explain. | 7.81 | 2.92 | 2.35 | 2.34 | 2.00 | |
| | | Deep L. | 4.76 | 1.47 | 1.00 | 0.80 | 0.81 | |
| | Synt. | Explain. | 12.00 | 5.51 | 4.15 | 3.77 | 3.12 | |
| | | Deep L. | 6.81 | 2.08 | 1.59 | 1.26 | 1.41 | |
| BiosecurID | Real | Explain. | 8.16 | 5.28 | 5.09 | 4.95 | 4.58 | |
| | | Deep L. | 9.85 | 6.11 | 5.58 | 5.55 | 4.16 | |
| | Synt. | Explain. | 8.33 | 5.70 | 5.16 | 5.58 | 4.41 | |
| | | Deep L. | 8.04 | 5.07 | 4.54 | 4.56 | 3.31 | |
| CEDAR | Real | Explain. | 5.19 | 3.64 | 1.99 | 1.85 | 1.25 | |
| | | Deep L. | 8.01 | 4.92 | 4.21 | 3.64 | 3.54 | |
| | Synt. | Explain. | 7.00 | 4.38 | 3.13 | 2.49 | 2.39 | |
| | | Deep L. | 4.58 | 1.58 | 1.31 | 0.94 | 1.01 | |
| MCYT | Real | Explain. | 21.17 | 18.24 | 15.93 | 13.79 | 10.85 | Skilled forgeries |
| | | Deep L. | 23.93 | 16.99 | 14.86 | 12.46 | 9.79 | |
| | Synt. | Explain. | 23.04 | 18.77 | 16.19 | 14.95 | 12.37 | |
| | | Deep L. | 22.42 | 16.28 | 12.63 | 10.59 | 8.72 | |
| Thai | Real | Explain. | 20.58 | 12.42 | 9.58 | 9.33 | 8.50 | |
| | | Deep L. | 20.25 | 10.25 | 7.75 | 8.08 | 7.75 | |
| | Synt. | Explain. | 24.25 | 14.83 | 12.42 | 12.17 | 11.08 | |
| | | Deep L. | 22.92 | 11.50 | 9.08 | 8.83 | 8.50 | |
| BiosecurID | Real | Explain. | 26.03 | 20.28 | 19.20 | 18.64 | 17.62 | |
| | | Deep L. | 31.08 | 22.30 | 19.96 | 20.59 | 18.19 | |
| | Synt. | Explain. | 25.27 | 19.52 | 18.89 | 18.38 | 17.31 | |
| | | Deep L. | 23.37 | 17.94 | 14.91 | 15.22 | 13.33 | |
| CEDAR | Real | Explain. | 12.12 | 7.20 | 4.70 | 4.24 | 3.71 | |
| | | Deep L. | 8.41 | 6.67 | 5.53 | 4.77 | 4.70 | |
| | Synt. | Explain. | 11.82 | 6.59 | 3.71 | 3.48 | 3.18 | |
| | | Deep L. | 5.15 | 2.95 | 2.27 | 1.52 | 1.59 | |

## 5.3. Studying the universal background model

In this subsection, we explored several aspects of the UBM. Once again, we use one signature as a reference, feature fusion at score level and the $\ell^1$ norm as understandable matching distance.

We collected the performance for both random and skilled forgeries by gradually increasing the number of genuine signatures in the UBM from 50 to 300. Figure 4 shows three subplots, one per dataset considered. Each subplot shows four solid lines for random and skilled forgery experiments, with explainable and DL features. Relatively stable performance is observed in all cases for morethan 100 different genuine signatures. Therefore, we can- not decide on a precise $n_{opt}$ number. In the following

*Table 6 State-of-the-art results in offline ASV*

| Method | Ref. | Performance (EER %) | | | |
|---|---|---|---|---|---|
| | | $th_{global}$ | | $th_{user}$ | |
| | | RF | SF | RF | SF |
| *Database: GPDS* | | | | | |
| CNN-BiLSTM [43] | FCV* | – | 10.16 | – | – |
| HOG, DRT ?DMML [67] | 5 | 2.15 | 20.94 | – | – |
| SigNet ?SVM [29] | 12 | – | – | – | 1.69 |
| Curvelet trans. ?OC-SVM [27] | 8 | – | 15.95 | – | – |
| Tangent Angle?SVM [8] | 12 | – | 14.82 | – | – |
| Poset-oriented grid ?SVM [77] | 5 | – | – | – | 9.87 |
| *Database: CEDAR* | | | | | |
| CNN-BiLSTM [43] | FCV* | – | 0 | – | – |
| on-2-off, SigCNN [35] | 5 | – | 6.41 | – | 4.50 |
| on-2-off, SigCNN [35] | 10 | – | 5.27 | – | 3.48 |
| CNN?Con. Loss Layer [71] | 5 | – | 2.50 | – | – |
| CNN?Con. Loss Layer [71] | 10 | – | 1.66 | – | – |
| Point-to-Set ?CNN [76] | 5 | – | 9.29 | – | 5.22 |
| Triplet Nets-Graph [46] | 10 | – | 12.27 | – | 5.91 |
| Archetypal analysis [78] | 5 | – | – | – | 2.07 |
| SigNet ?SVM [29] | 8 | – | – | – | 4.77 |
| Poset-oriented grid ?SVM [77] | 5 | – | – | – | 4.12 |
| Curvelet trans. ?OC-SVM [27] | 8 | – | 7.83 | – | – |
| DCNN ?PDSN [38] | 5 | – | 7.40 | – | 4.37 |
| Waveforms?GP [65] | 8 | – | 8.71 | – | – |
| Meta-learning [30] | 8 | – | 10.21 | – | 7.07 |
| *This work: explainable ASV* | 1 | 5.19 | 12.12 | – | – |
| *This work: explainable ASV* | 5 | 1.99 | 4.70 | – | – |
| *This work: explainable ASV* | 10 | 1.25 | 3.71 | – | – |
| *Database: MCYT* | | | | | |
| On-2-off, SigCNN [35] | 5 | – | 5.82 | – | 3.42 |
| On-2-off, SigCNN [35] | 10 | – | 4.55 | – | 2.01 |
| CNN?Con. Loss Layer Tsourounis et al. [71] | 5 | – | 2.61 | – | – |
| CNN?Con. Loss Layer Tsourounis et al. [71] | 10 | – | 1.62 | – | – |
| HOG, DRT?DMML [67] | 5 | 1.73 | 13.44 | – | – |
| Poset-oriented grid ?SVM [77] | 5 | – | – | – | 6.02 |
| Archetypal analysis [78] | 5 | – | – | – | 3.97 |
| SigNet ?SVM [29] | 10 | – | – | – | 2.87 |
| VLAD ?KAZE [54] | 10 | – | 5.60 | – | – |
| DCNN ?PDSN [38] | 5 | – | 7.12 | – | 3.78 |
| Point-to-Set ?CNN [76] | 5 | – | 9.21 | – | 4.86 |
| Triplet Nets-Graph [46] | 10 | – | 9.16 | – | 3.91 |
| Waveforms?GP [65] | 10 | – | 7.55 | – | – |
| Meta-learning [30] | 5 | – | 15.37 | – | 12.77 |
| Handcrafted ?DTW [18] | 1 | 6.79 | 19.71 | – | – |
| *This work: Explainable ASV* | 1 | 6.29 | 21.17 | – | – |
| *This work: Explainable ASV* | 5 | 3.86 | 15.93 | – | – |
| *This work: Explainable ASV* | 10 | 1.80 | 10.85 | – | – |
| *Database: Thai* | | | | | |
| SCUT ?CNN [13] | 5 | 0.19 | 7.10 | – | – |

*Table 6 (continued)*

| Method | Ref. | Performance (EER %) | | | |
|--------|------|---------------------|---|---|---|
| | | $th_{global}$ | | $th_{user}$ | |
| | | RF | SF | RF | SF |
| LTP ?oBIFs [13] | 5 | 1.09 | 10.91 | – | – |
| ERL [13] | 5 | 3.02 | 17.80 | – | – |
| Textural ?HMM [12] | 5 | 2.01 | 11.08 | – | – |
| *This work: Explainable ASV* | 1 | 7.81 | 20.58 | – | – |
| *This work: Explainable ASV* | 5 | 2.35 | 9.58 | – | – |
| *Database: BiosecurID* | | | | | |
| LBP ?SVM ?Attributes [53] | 4 | 1.66 | 15.55 | – | – |
| LBP ?SVM [24] | 4 | 4.81 | 20.28 | – | – |
| Handcrafted ?DTW [18] | 1 | 10.65 | 25.91 | – | – |
| *This work: explainable ASV* | 1 | 8.16 | 26.03 | – | – |
| *This work: explainable ASV* | 5 | 5.09 | 19.20 | – | – |

*\*FCV Fold cross validation process*

experiments, we set $n_{opt}$ =300, as we did in our previous work [18]. Additionally, we see that the curves for random and skilled forgeries are slightly closer together. In these cases, we also see that DL features do not always guaranteebetter performance, especially for skilled forgeries.

It is worth noting that the size of the UBM can beincreased, which may be advantageous for computer vision applications. However, the objective of our explainableoffline ASV is to serve as an automated tool for FHE who are accustomed to designing manual UBM. Therefore, a size of 300 signatures could be deemed large for FHE, but it would be considered a standard UBM size for computer vision purposes.

Also, we included fake signatures in the UBM used for matching the genuine one. As such, the UBM was enlarged   to $2 \cdot n_{opt}$ offline specimens. In general, we observed that adding forgeries in the UBM barely improves performance. In MCYT or Thai with explainable features, the results were worsened by less than 1. No relevant improvements were seen in RF with BiosecurID or Thai when DL features were used. As for the SF, BiosecurID constantly improves, whereas no effect is observed in Thai, and MCYT is slightly worsened. As for an easy-to-explain system in a court, we avoid complicating the UBM by introducing forgeries. Hence, this analysis suggests using only genuine signatures in the UBM.

Data privacy helps in the design of a confidentiality system, which is another goal considered in the explicability of the algorithms [5]. Accordingly, we designed a UBM with artificial specimens as a further option for our system. To this end, we randomly chose 300 different signatures from the first identities in a synthetic offline database [22]. All experiments were repeated using this synthetic UBM. To analyze the use of this UBM, we reportthe results obtained in Table 4, when a real UBM was usedunder the same conditions. Overall, we can observe that theperformance does not change significantly due to the effectof the real versus synthetic UBM. On the contrary, it sometimes improves, as in the case of BiosecurID, or is slightly worse (see results with Thai).

In some applications in biometrics, synthetic signatures alleviate conflicts with data protection regulations and copewith insufficient training data. However, even though privacy is one of the requirements in explainable systems [5], some applications prefer real signatures in the UBM. According to this dichotomy, we analyze below the pro- posed explainable ASV, thus offering results with a real and a synthetic UBM.

*5.4. Effect of using multiple reference signatures*

We further compared the performance with explainableand DL features when different reference signatures were used. Commonly, the more knowledge a system has, the less the error will be. For this experiment, we used the best configurations found fused features and $\ell^1$ norm. In addition to the LR and probabilities of belonging to the UBM, we computed the probability of belonging to the reference set when more than one signature is used. These latter probabilities are fused at the score level by using the same weights already found.

We can see in Table 5 a consistent behavior of the proposed system since the more reference signatures there are, the better the performance. This means that we obtaina better representation of $P(R)$. As expected, the best results were obtained using DL features. Aside from BiosecurID, real signatures in the UBM seem to be a better option for performance. It should be considered therefore that interpretability was successful, despite the performance loss since it leads to explainable ASVs. This experiment allows us to compare our results with the state-of-the-art.

*5.5. Comparative analysis with the state of the art*

The proposed explainable ASV is now compared to previous work. We aim to determine whether the achieved performance using our explainable system is adequate for practical use. Indeed, one of the most challenging tasks is to evaluate fairly the state-of-the-art results in offline signature verification. Despite recent efforts with the Thai dataset [13], themain reason is the lack of standard benchmarks or com- petitions fixing experimental protocols and metrics for ASV evaluation.

Despite this difficulty, Table 6 tries to overview results in ASV using several publicly available popular databases. Other complications to reasonably analyzing prior literature are using global versus custom thresholds in the system or experimenting with random or skilled forgeries.Moreover, evaluation metrics like accuracy, FAR, and FRRare commonly used throughout the literature. For simplicity and tradition in the field [16], Table shows some works that use the EER and have a performance which is competitive.

More importantly, in this work, the contributions in Table 6 are based on machine learning (ML) and DL techniques. According to [1, 28], DL is at the extreme of unexplainable systems, and some ML techniques could be considered easier to explain, such as random forests (RFs), which are not commonly used in ASV because of theirpoor performance. Consequently, the use of the challenging systems by the FHE in a courtroom cannot be guar- anteed because their results cannot be easilyexplained [44].

To this end, using a fully explainable system for automatically verifying signatures would lead to a loss of performance. This work can quantify such a performance loss by comparing our results with the state-of-the-art in Table 6. Our most explainable configuration is based on explainable features, '$^1$ norm as matching distance, real signatures in the UBM, and using a global threshold for evaluation. We estimate a weak performance of 2.16 and 2.48 perceptual points for RF and SF, respectively, for the most challenging system in the Thai database. Nevertheless, it is worth pointing out that our system outperformed the results given in [12]. In the case of BiosecurID, we lost about 3.62 and 4.73 of performance for RF and SF. This is compared with [53], which used four signatures as references and was the best performing system in Table. Regarding the work proposed in [67], we quantify a performance loss of our system of 2.13 and 2.49 for RF and SF, respectively, with the MCYT corpus. In summary, evenif we lose some performance, the advantage of offering explicable ASVs validates the use of our system in specificapplications, where this characteristic is critical, such as in forensic applications.

## 6. Conclusion

In this paper, we proposed a novel explainable offline signature verifier to support FHEs. We introduced a universal background model of signatures from third-party signers to improve the accuracy of our system. To preserve privacy, we added synthetic signatures to the UBM. We also considered explainable features and understandable distance matching.

Our ASV can provide objective evidence of whether a signature is genuine or not in terms of likelihood ratios and probabilities of belonging to the UBM and the reference set. This makes it suitable for use in forensic settings, where it is important to be able to explain the decisions made by the ASV.

Our experiments demonstrated that our ASV can achieve competitive performance in an explainable setting. We also showed that explainable features and an understandable distance matching based on the '$^1$ norm can be used to maintain a state-of-the-art performance level. This suggests that handcrafted features should still be considered for ASV applications where system explainability is crucial. We believe that this research will help to narrow the gap between the forensic and pattern recognition communities by providing a novel and explainable offline signature verifier.